%% file: 0_main.tex
\documentclass[10pt,journal]{IEEEtran}
\IEEEoverridecommandlockouts

\usepackage{amsmath,amssymb,amsfonts}
\usepackage{graphicx}
\usepackage{textcomp}
\usepackage{xcolor}
\usepackage{amsmath} 
\usepackage{amssymb}  
\usepackage{bm}
\usepackage{mathtools}
\usepackage{booktabs}
\usepackage{float}
\usepackage{placeins}
\usepackage{url}
\usepackage{algpseudocode}
\usepackage{algorithm}
\usepackage{multirow}
\usepackage{xcolor}
\usepackage{makecell}
\usepackage{threeparttable}
\usepackage{wrapfig}
\usepackage{cuted}
\usepackage{capt-of}
\usepackage{balance}
\makeatletter
\let\NAT@parse\undefined
\makeatother
\usepackage[colorlinks=true, citecolor=blue, linkcolor=blue, urlcolor=blue]{hyperref}
\def\BibTeX{{\rm B\kern-.05em{\sc i\kern-.025em b}\kern-.08em
    T\kern-.1667em\lower.7ex\hbox{E}\kern-.125emX}}

\DeclareMathOperator{\eventually}{\Diamond}
\DeclareMathOperator{\always}{\Box}
\DeclareMathAlphabet{\mathmybb}{U}{bbold}{m}{n}

\begin{document}

\title{\LARGE \bf Learning Robot Policies from Sparse Success Signals via\\ STL-Guided Stein Variational Policy Gradient
}

\author{
Hongrui Zheng$^{1}$, Cristian Ioan Vasile$^{2}$, Antonio Loquercio$^{\dagger,1}$, Rahul Mangharam$^{\dagger,1}$
\thanks{$^\dagger$Equal Advising. $^{1}$Department of Electrical and Systems Engineering, University of Pennsylvania, Philadelphia, PA, USA. $^{2}$Mechanical Engineering and Mechanics Department, Lehigh University, Bethlehem, PA, USA. Correspond to \tt\small hongruiz@engineering.upenn.edu.}
}

\maketitle

\begin{strip}
\vspace{-74pt}
    \centering
    \includegraphics[width=\textwidth]{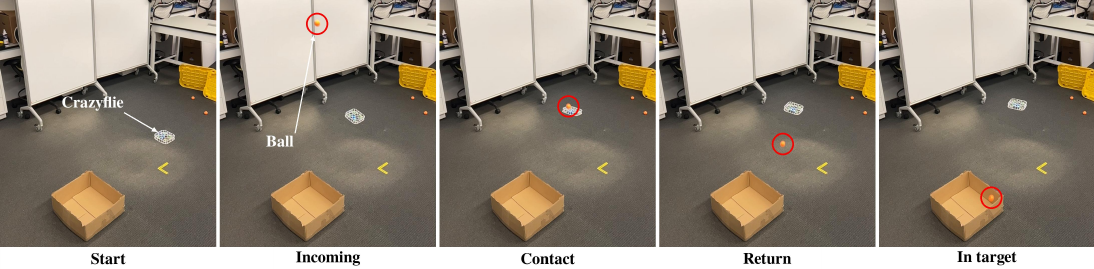}
    \captionof{figure}{STL-SVPG combines Signal Temporal Logic (STL) robustness gradients and Stein Variational Gradient Descent (SVGD) to train low-level control policies in tasks with very sparse success signals, such as the drone's ping-pong task shown above. Shown is a successful real-world rollout of a policy trained in simulation with our approach. From left to right: initial hover, incoming ball, contact, ball return, and in target. Red circles highlight the ball.}
    \label{fig:real-ping-pong}
    \vspace{-10pt}
\end{strip}

\input{1_abstract}
\input{2_intro}
\input{3_method}
\input{4_exp}
\input{5_conclude}

\balance
\bibliographystyle{IEEEtran}
\bibliography{references}

\end{document}

%% file: 1_abstract.tex
\begin{abstract}
Learning robot policies for tasks with sparse success signals is challenging when completion depends on coordinated actions, precise contact outcomes, or satisfying several conditions together.
Intricate physical interactions with the world further complicate these requirements. Prior work using conventional reward shaping mechanisms provides dense feedback but local progress might not translate into eventual task completion.
We present Signal Temporal Logic-guided Stein Variational Policy Gradient (STL-SVPG), a population-based method that uses smooth STL robustness as a trajectory-level training objective. Differentiating this objective through the dynamics assigns credit to policy actions according to their effect on the complete task specification, rather than local progress alone.
We evaluate the approach on six quadcopter and manipulator tasks that involves event-triggered responses, strictly ordered behavior, responses within specified deadlines, and physical interaction with the world. STL-SVPG achieves the highest mean success rate among the compared methods on five of six benchmarks. Simulation-trained policies trained in simulation transfer temporal and contact task behavior to the real world.
\end{abstract}

%% file: 2_intro.tex
\section{Introduction}

Learning low-level robot control policies is challenging when task success signals are extremely sparse. Exploration may produce many unsuccessful trajectories before encountering a successful one, leaving little feedback to distinguish useful behavior or identify which actions matter. For example, a robot may need to coordinate several movements before completing an assembly, or strike an object so that it reaches a distant target. Ordering, timing, and simultaneous constraints can further restrict the set of successful trajectories.

Reward shaping~\cite{ng_policy_1999} and model-based reinforcement learning are two common ways to assist learning in these settings. Handcrafted or automatically designed dense rewards~\cite{ma_eureka_2024} provide intermediate guidance, but require choices about which behaviors to encourage. Designing informative dense rewards is an iterative engineering process: changes to the task can require redesign and repeated training and evaluation. Automated reward design reduces manual effort but still relies on this feedback loop.
Learned models in RL~\cite{janner_when_2019} can support lookahead and policy improvement with predicted experience, but do not by themselves provide informative task feedback. When useful outcomes require long action sequences, short planning horizons may miss them, while prediction errors can compound over longer model rollouts.

Therefore, for these sparse-completion tasks, informative feedback from unsuccessful trajectories is essential to practical learning efficiency: otherwise, policy improvement must wait for rare successful exploration.
Signal Temporal Logic (STL)~\cite{maler_monitoring_2004} does exactly this by specifying logical and temporal requirements over real-valued signals. The robustness semantics~\cite{donze2010robust,fainekos2009robustness} is a quantitative measure that assign the margin to success to a trajectory instead of a binary signal.

However, optimizing the robustness objectives through robot dynamics remains challenging: while smooth robustness makes gradients available, they remain concentrated on a few critical steps, vanishing over long trajectories~\cite{pant_smooth_2017,leung2023backpropagation,kapoor_stlcg_2025}.
STL-SVPIO~\cite{zheng_stl-svpio_2026} addresses these challenges by applying Stein variational optimization through differentiable STL robustness to optimize control sequences in nonlinear dynamical systems. 
However, similarly to prior work in STL planning~\cite{raman_reactive_2015,pant_smooth_2017}, it only finds open-loop control sequences from fixed initial conditions, with long computation time that makes online re-planning infeasible.

To amortize this computation, prior work trains a feedback policy using STL robustness as a reward~\cite{li_reinforcement_2017,meng2023signal}.
However, this leaves temporal credit assignment to exploration, while differentiable predictive controllers still rely on online trajectory prediction and backup mechanisms. These limitations motivate learning low-level feedback policies by differentiating trajectory-level robustness, where unsuccessful rollouts are still valuable training signals.

We propose STL-SVPG to learn low-level control policies for tasks with sparse success signals by exploiting an explicit STL specification of task completion. 
We use smooth STL robustness to quantify how trajectories satisfy or violate the specification. In contrast to reward engineering, the STL specification generates quantitative feedback from the \emph{same} temporal requirements used to judge completion, reducing the need to design sub-task rewards.
Like prior STL-based approaches for policy optimization~\cite{li_reinforcement_2017,meng2023signal}, we derive learning feedback from quantitative satisfaction of the task specification. In contrast to prior work, however, our approach couples trajectory-level robustness gradients with Stein Variational Gradient Descent (SVGD)~\cite{liu_stein_2016,liu2017stein}, resulting in a population of feedback policies that interact during learning: kernel-weighted updates share improvement directions between similar policies, while repulsive forces encourage diversity of behaviors. This strategy couples exploitation and exploration during optimization.

The goal of this paper is to leverage STL task specifications' ability to provide improvement signal even in unsuccessful rollouts to train low-level control policies for robots in tasks with sparse success signals. Empirically, our framework achieves the highest mean success among a set of reinforcement learning baselines with shaped dense rewards on five of six benchmarks. In addition, the policies trained with STL-SVPG in simulation transfer zero-shot to physical hardware. Overall, our contributions are as follows:
\begin{enumerate}
    \item We introduce STL-SVPG, a specification driven policy learning framework that turns overall task requirements into actionable policy gradients for low-level feedback control even from unsuccessful trajectories. 
    \item A policy optimization method that shares information amongst a population of policy parameters that combines task-directed improvements with population-based exploration.
    \item We investigate what makes this learning framework effective: how STL robustness gradients guide improvement and how population interactions expand exploration for better policies.
    \item Validation of simulation-only trained STL-SVPG policies on two tasks in the real world.
\end{enumerate}


\section{Related Work}
\label{sec:related}
Learning with sparse task-success signals can be assisted by shaped rewards, explicit task-progress representations, or quantitative specifications. We discuss existing work in each of these directions to give readers a better context for our contribution.

\textbf{Reward shaping:}
Potential based reward shaping~\cite{ng_policy_1999} is the foundational work, it characterizes reward transformations that preserve optimal policies under their MDP assumptions.
More recent advances in reward design with coding agents~\cite{ma_eureka_2024} generate and iteratively improve reward code using large language models and policy-training feedback. The dense reward for each task we used in our experiments is also defined using this approach.
An adjacent approach is Hindsight Experience Replay (HER) ~\cite{andrychowicz_hindsight_2017}. It relabels achieved goals to learn from unsuccessful attempts rather than dense reward shaping.
In contrast, our approach utilizes a smooth robustness surrogate of an STL specification, and we use automatic differentiation to generate policy gradients. This provides task-directed feedback before any success is even discovered, while reward shaping still requires successful or close to successful exploration, and HER turns alternative goals into successes.

\textbf{Task-progress representations:}
Reward machines (RMs)~\cite{icarte_using_2018,icarte_reward_2022} represent task progression through finite-state representations. RMs can be decomposed into subproblems that are learned with off-policy Q-learning. Although a powerful approach for task decomposition, their use for training robotic policies~\cite{camacho_disentangled_2020} is still limited.
LTL2Action~\cite{vaezipoor_ltl2action_2021} exploits LTL progression, which transforms the original non-Markovian problem into a Taskable MDP, then teaches agents to follow LTL-based instructions.
Our approach does not need explicit decomposition of the original task. The STL specifications represent progression through different predicate functions and logical and temporal operators.

\textbf{Learning with quantitative specifications:} In the early years, Li et al. ~\cite{li_reinforcement_2017} has explored learning policies using the robustness of truncated linear temporal logic as a reward in RL.
Varnai and Dimarogonas~\cite{varnai_prescribed_2019,dimos_learning_2019} also proposed a general RL framework under STL signals.
These works either used simplistic tasks or the benchmarks were limited to point-mass systems.
Later, Saxena et al.~\cite{saxena_funnel-based_2024} convert STL requirements into time-dependent shaped rewards and learns policies, using TD3 for continuous-action experiments.
More recently, TGPO~\cite{meng_tgpo_2025} has decomposed STL tasks into timed subgoals and invariant constraints, combining temporal search with policy learning.
On the other hand, Smooth Operator~\cite{pant_smooth_2017} and STLCG(++)~\cite{leung2023backpropagation,kapoor_stlcg_2025} make STL specifications accessible to automatic differentiation.
STLNPC~\cite{meng2023signal} also uses this gradient. It trains a neural controller by maximizing predicted-trajectory robustness and uses predictive deployment with a backup policy.
As we will show in the problem formulation, our population based method maximizes differential entropy of the learned policy distribution, which ends up bringing diversity in exploration.
We will also show in comparison that our method is superior to using gradient descent alone.

%% file: 3_method.tex
\section{Background}\label{sec:background}

\subsection{Signal Temporal Logic (STL)}\label{sec:background_stl}
\textbf{Syntax:}
Signal Temporal Logic (STL)~\cite{maler_monitoring_2004} provides a compact representation for imposing spatiotemporal requirements over trajectories $s$. For example, we can express ``Reach a goal within five seconds while remaining in a safe region throughout the episode'' in a compact representation: $\phi=\eventually_{[0,5]}\mu_{\mathrm{goal}}\wedge\always_{[0,5]}\mu_{\mathrm{safe}}$, where $\mu_{\mathrm{goal}}$ and $\mu_{\mathrm{safe}}$ are predicate functions that measures whether the robot is in the goal, and being safe.
In this work, we use STL formulae to express robotic task objectives.
The STL syntax is
\begin{equation}\label{eq:stl_syntax}
  \phi := \top \mid \mu^g \mid \neg\phi \mid \phi_1\wedge \phi_2 \mid \phi_1\,\mathbf{U}_{[t_1,t_2]}\,\phi_2,
\end{equation}
where $\top$ is the truth value, $\mu^g:=g(s(t))\ge 0$ is an atomic predicate, the value of the function $g:\mathbb{R}^n\rightarrow\mathbb{R}$ determines the truth value of the predicate.
Boolean operators $\wedge$ and $\neg$ denotes ``conjunction" and ``negation" respectively.
Temporal operator $\mathbf{U}$ denotes ``until": formula $\phi_1$ has to hold until $\phi_2$ holds in the time interval $[t_1,t_2]$.
We can then use these primitives to define other operators.
For example, disjunction is defined as $\phi_1 \vee \phi_2 := \neg(\neg\phi_1 \wedge \neg\phi_2)$;
``Eventually" is defined as $\eventually_{[t_1,t_2]}\phi := \top\,\mathbf{U}_{[t_1,t_2]}\,\phi$;
``Always" is defined as $\always_{[t_1,t_2]}\phi := \neg\eventually_{[t_1,t_2]}\neg\phi$; 
And ``Implication" is defined as $\phi_1\implies\phi_2:=\neg\phi_1\vee\phi_2$.

\textbf{Quantitative Semantics:}
STL uses \textbf{robustness} $\rho^\phi(s,t)\in\mathbb{R}$ to measure how strongly a given trajectory satisfies or violates a specification $\phi$. $\rho>0$ implies satisfaction and $\rho<0$ implies violation. 
Each predicate gives a signed satisfaction margin. Conjunction and always use the smallest relevant margin, whereas disjunction and eventually use the largest. Composing these operations yields one trajectory-level score whose sign indicates satisfaction or violation. We use the standard robustness semantics~\cite{donze2010robust,fainekos2009robustness} in this work.

\textbf{Differentiable STL:}
Smooth max and min approximators~\cite{pant_smooth_2017} and parsing robustness evaluation into computation graphs~\cite{leung2023backpropagation,kapoor_stlcg_2025} have enabled automatic differentiation of the robustness value of a trajectory to the states in the trajectory. In conjunction with differentiable dynamics, we can compute gradients directly from robustness to the policy's actions.


\subsection{Stein Variational Gradient Descent}

SVGD~\cite{liu_stein_2016} performs gradient descent by transporting a set of particles in the parameter space to match the target posterior representing the optimal objective.
At each iteration $m$, SVGD updates the particles by $\theta_i^{m+1}=\theta_i^m+\epsilon\varphi^*(\theta_i^m)$, where $\{\theta^0_i\}_{i=1}^N$ is the set of initial particles, $\epsilon$ is the step size, and the function $\varphi^*(\cdot)$ denotes optimal perturbation that maximally decreases the KL-divergence between the proposal distribution and the target posterior:
\begin{equation}\label{eq:stein_dir}
\begin{aligned}
\varphi&^*(\theta_i)=\\&\frac{1}{N}\sum_{j=1}^N\bigg[ \underbrace{K(\theta_j, \theta_i) \nabla_{\theta_j} \log p(\theta_j)}_{\text{Attractive Force}} 
   + \underbrace{\nabla_{\theta_j} K(\theta_j, \theta_i)}_{\text{Repulsive Force}} \bigg].
\end{aligned}
\end{equation}
The two portions of the update directions represent the attraction force on the particles towards areas of higher expected score, with the kernel's value enabling ``information sharing" between particles, and the repulsion force preventing mode collapse. 

\section{Problem Formulation}
We formulate task-directed policy learning as variational inference (VI) over policy parameters. Let $\pi_\theta$ be a feedback policy and $\tau_\theta$ its induced finite-horizon trajectory, and $\xi\sim\mathcal{D}$ be randomized initial conditions and task inputs. With a task STL specification $\phi$, the policy objective is the expected robustness:
\begin{equation}
    J(\theta)=\mathbb{E}_{\xi,\pi_\theta}
    [\rho_\phi(\tau_\theta)].
\end{equation}
As in prior work for policy optimization with SVGD~\cite{ziebart2008maximum,liu2017stein,lei2026reinforcement}, we  define the \emph{target distribution} over policy parameters as the task-conditioned Gibbs density. The intuition behind this choice is that this distribution gives density to policies proportional to their expected return: the higher their Q function, the greater the density. By approximating this target distribution, we naturally find good performing policies. And by replacing the expected return with expected STL robustness, this gives us:

\begin{equation}\label{eq:policy_gibbs_target}
\begin{aligned}
    p^\star(\theta)&=Z^{-1}
        \exp\!\left(J(\theta)/\alpha\right),\\
    Z&=\int \exp\!\left(J(\theta)/\alpha\right)d\theta.
\end{aligned}
\end{equation}
where $\alpha>0$ is the temperature controlling the sharpness of the target distribution and $0<Z<\infty$ is a normalizing factor that makes it a valid distribution. Let $q(\theta)$ denote the \emph{variational approximation} to this target. The VI objective is
\begin{equation}\label{eq:policy_distribution_objective}
    \min_q\;\mathrm{KL}(q\|p^\star),
\end{equation}
over probability densities for which the objective is well-defined. By substituting $\log p^\star=J/\alpha-\log Z$ into the KL-divergence definition, we can minimize the \emph{variational free energy}
\begin{equation}\label{eq:policy_free_energy}
\begin{aligned}
    \mathcal{F}(q) &= \alpha\,\mathrm{KL}(q\|p^\star)-\alpha\log Z \\
    &= -\mathbb{E}_{\theta\sim q}[J(\theta)]-\alpha\mathcal{H}(q),
\end{aligned}
\end{equation}
where $\mathcal{H}(q)=-\int q(\theta)\log q(\theta)\,d\theta$ is the differential entropy in policy parameter space. With these two terms, minimizing free energy maximizes both expected robustness and entropy, hence balancing exploitation and exploration. The target $p^\star$ is optimal for this entropy-regularized objective. Note that the entropy here is in the policy parameter space, and not the usual policy action space.
We use SVGD~\cite{liu_stein_2016} to transport policy particles representing $q$ toward $p^\star$.
In the next section, we discuss how we adapt this formulation to optimize a policy with an STL robustness score.

\section{STL-Guided Stein Variational Policy Gradient}
\label{sec:stl_svpg}

Building on the target distribution in Equation~\eqref{eq:policy_gibbs_target}, we represent the variational approximation $q$ by a population of policy parameter particles $\Theta=\{\theta_i\}_{i=1}^{N}$.
We consider a finite-horizon, discrete-time dynamical system
\begin{equation*}
    x_{t+1}=f(x_t,u_t), \qquad u_t\sim\pi_\theta(o_t),
\end{equation*}
where $x_t$, $o_t$, and $u_t$ are the state, policy observation, and control input. Initial conditions and task inputs are sampled through $\xi\sim\mathcal{D}$, as in Section IV. A rollout is $\tau_\theta=(x_0,o_0,u_0,\ldots,x_H)$.
For optimization, we replace the exact robustness in $J$ with a differentiable surrogate $\widetilde{\rho}_\phi$ by using LogSumExp~\cite{pant_smooth_2017,leung2023backpropagation,kapoor_stlcg_2025} as the smooth maximum operator:
\begin{equation}\label{eq:smooth_policy_objective}
    \widetilde{J}(\theta)
    =\mathbb{E}_{\xi,\pi_\theta}
        [\widetilde{\rho}_\phi(\tau_\theta)].
\end{equation}
For $n$ scalar inputs $z_1,\ldots,z_n$, we define the smooth maximum and minimum as
\begin{equation}
\begin{aligned}
    \operatorname{smax}_{\beta}(z_1,\ldots,z_n)
    &= \frac{1}{\beta}\log\sum_{i=1}^{n}\exp(\beta z_i),\\
    \operatorname{smin}_{\beta}(z_1,\ldots,z_n)
    &= -\operatorname{smax}_{\beta}(-z_1,\ldots,-z_n),
\end{aligned}
\end{equation}
where $\beta>0$ controls the approximation sharpness and is distinct from the Gibbs temperature $\alpha$. Larger $\beta$ yields a closer approximation to the exact maximum and minimum, while smaller $\beta$ distributes gradient weight more broadly across the inputs. 
We use automatic differentiated simulation implemented with MuJoCo MJX~\cite{todorov2012mujoco} in JAX~\cite{jax2018github} with soft contact models to propagate robustness gradients through the trajectory to the controls. We use smoothed robustness only for training but not for evaluation, which is done via exact robustness.

Following SVGD~\cite{liu_stein_2016} and SVPG~\cite{liu2017stein}, we derive the policy gradient. For the Gibbs target in Equation~\eqref{eq:policy_gibbs_target} in Section IV, the log-density gradient required by SVGD is the expected robustness gradient scaled by $(1/\alpha)$: $\nabla_\theta\log p^\star(\theta)=\alpha^{-1}\nabla_\theta J(\theta)$. We therefore obtain the policy update by substituting this scaled robustness gradient into Equation \eqref{eq:stein_dir}.
Thus, the Stein direction for policy optimization is:
\begin{equation}\label{eq:stl_policy_stein}
    \varphi_i=\frac{1}{N}\sum_{j=1}^{N}\bigg[K(\theta_j,\theta_i)\frac{\nabla_{\theta_j}\widetilde{J}(\theta_j)}{\alpha}+\nabla_{\theta_j}K(\theta_j,\theta_i)\bigg].
\end{equation}
The kernel-weighted score shares task-improving directions across particles, while the kernel derivative provides repulsion in parameter space. 

Differentiating the smooth robustness $\widetilde{\rho}_\phi$ through the simulated dynamics yields action gradients that can guide improvement even when a rollout does not satisfy the specification yet.
We form the local policy-gradient estimate $\widehat g_i$ by multiplying these action gradients by the policy's action Jacobians, summing over time, and averaging over $E$ rollouts. 

Collected observations and sampled policy noise are held constant in this calculation. Thus, $\widehat g_i$ is a local approximation rather than the full closed-loop policy derivative. Storing the action gradients allows their contributions to be accumulated through randomized transition minibatches without repeatedly differentiating the simulator.

To measure behavioral differences between policies, we compare their actions on a shared, randomly sampled observation batch $\mathcal O$, rather than relying on parameter distances, which doesn't necessarily reflect behavioral similarity~\cite{peters2008reinforcement}. With $u_\theta(\mathcal O)$ denoting the concatenated policy actions, we use
\[
K(\theta_j,\theta_i)
=\exp\!\left(
-\frac{\|u_{\theta_j}(\mathcal O)-u_{\theta_i}(\mathcal O)\|^2}{h}
\right),
\]
where $h>0$ is the bandwidth.
We summarize the training procedure in Algorithm~\ref{alg:stl_svpg}.



\begin{algorithm}[t]
\caption{STL-SVPG}
\label{alg:stl_svpg}
\begin{algorithmic}[1]
\Require STL formula $\phi$, initial particles of policy weights $\{\theta^0_i\}_{i=1}^{N}$, $E$ rollouts per particle, horizon $H$, iterations $M$, initial condition randomization $\mathcal{D}$, step size $\epsilon$, temperature $\alpha$, kernel $K$
\For{$m=1,\ldots,M$}    
    \For{$i=1,\ldots,N$} \Comment{In parallel}
        \State Collect rollouts $\{\tau_{i,e}\}_{e=1}^{E}$
               under $\pi_{\theta_i}$ and $\xi_{i,e}\sim\mathcal D$
        \State Differentiate smooth trajectory robustness through
               the simulator and store action gradient with transitions:
               \[
               d_{i,e,t} = \frac{\partial \widetilde{\rho}_\phi(\tau_{i,e})}{\partial u_{i,e,t}}
               \]
        \State Calculate policy action Jacobians:
                \[
                J_{i,e,t}
                =
                \left.
                \frac{\partial u_{i,e,t}}
                {\partial\theta}
                \right|_{\theta=\theta_i}
                \]
        \State Accumulate $J_{i,e,t}^{\top}d_{i,e,t}$ over
               randomized transition minibatches to get the local policy gradients:
        \[
            \widehat g_i
            =\frac{1}{E}\sum_{e=1}^{E}\sum_{t=0}^{H-1}
              J_{i,e,t}^{\top}d_{i,e,t}
        \]
    \EndFor
    \State Sample shared kernel observations $\mathcal O$
        \State Compute kernel with shared observation action similarities
        \[K(\theta_j,\theta_i)
            =\exp\!\left(
                -\frac{\|u_{\theta_j}(\mathcal{O})
                 -u_{\theta_i}(\mathcal{O})\|^2}{h}
            \right)
        \]
        \State Compute Stein direction
        \[
            \varphi_i =
            \frac{1}{N}
            \sum_{j=1}^{N}
            \left[
                K(\theta_j,\theta_i) \widehat{g_j}/\alpha
                +
                \nabla_{\theta_j} K(\theta_j,\theta_i)
            \right]
        \]
        \State Simultaneously update $\theta_i\gets\theta_i+\epsilon\varphi_i$ for all $i$
\EndFor
\State \Return policy population; select a policy on separate validation cases
\end{algorithmic}
\end{algorithm}

%% file: 4_exp.tex
\section{Experimental Setup}
\label{sec:exp_setup}

\begin{figure*}[t]
    \centering
    \includegraphics[width=0.9\linewidth]{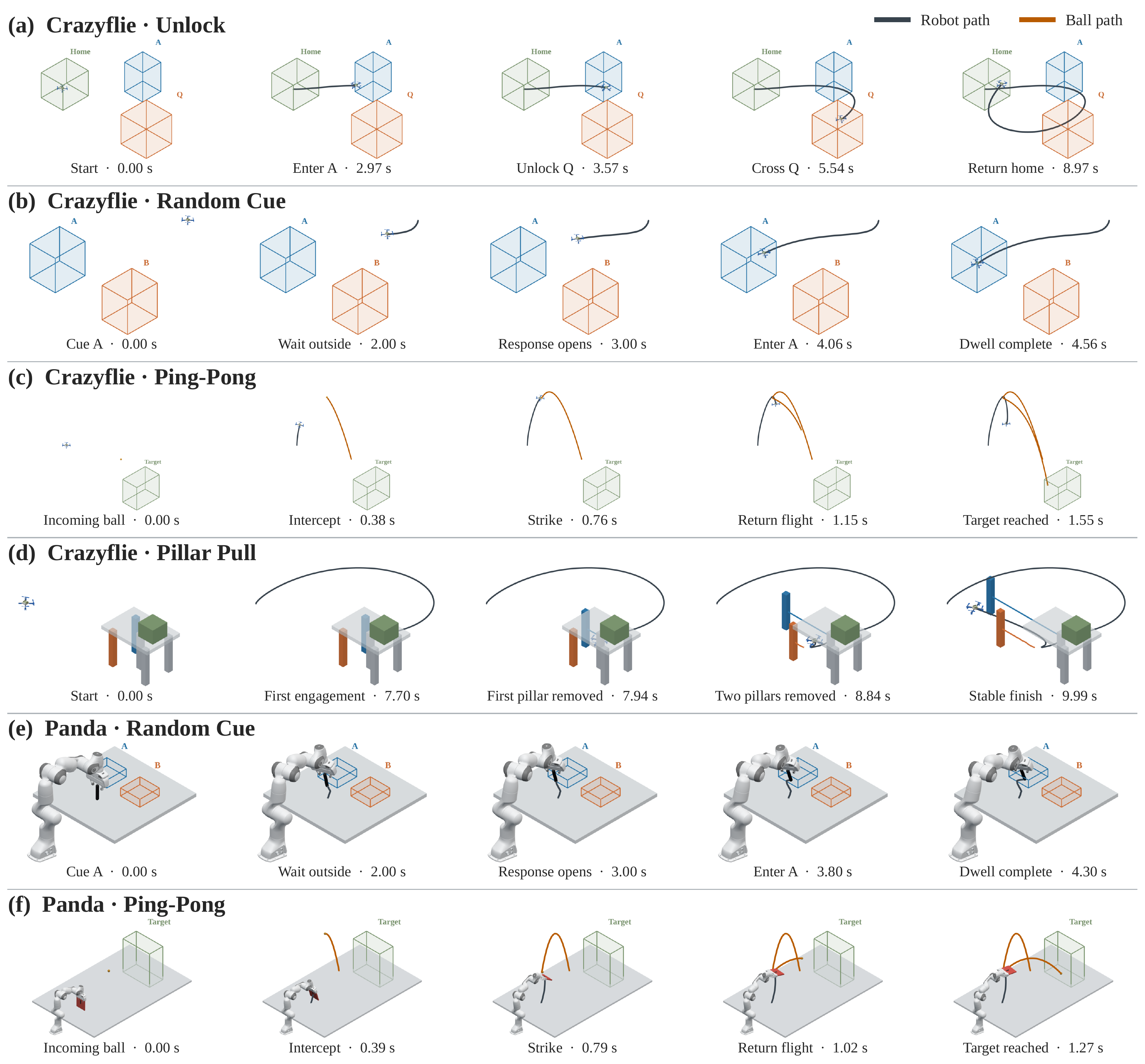}
    \caption{Example of successful STL-SVPG rollout in all tasks. In Unlock and both Random Cue tasks, the policy learned stable control that satisfies the timing and order requirements. In Crazyflie and Panda Ping-pong, it learned the ball return strikes reliably. And finally in Pillar Pull, it learned to bump two pillars without disturbing the balance of the plate and the block, without specifying which two pillars to be removed.}
    \label{fig:pillar_pull_rollout}
    \vspace{-10pt}
\end{figure*}

In the experiments, we select sparse success signal tasks on two different embodiments. These tasks combine continuous control with requirements for ordered actions, event-triggered responses, and contact interaction based outcomes.
The chosen scenarios test whether the algorithm can learn successful physical behavior while also satisfying the logical and temporal specifications that result in extremely sparse success signals.
We mainly test our learning algorithm on the Crazyflie~\cite{noauthor_crazyflie_nodate} platform, but also show the same algorithm generalizing to a different embodiment on a Franka Emika Panda manipulator.

\subsection{Robot Simulations}
We simulate Crazyflie rigid-body dynamics in MuJoCo MJX~\cite{todorov2012mujoco}, using a cascaded body-rate PID controller, first-order motor dynamics, and rotor-dependent aerodynamic drag used in~\cite{pasumarti_agile_2025}. Actions in the simulation are normalized collective thrust and body rate commands expected at 100\,Hz while control and physics run at 500\,Hz.

The Panda environment uses the seven-joint model for forward kinematics and bounded joint velocity control provided by MuJoCo. Joint positions are integrated at 5\,ms and an action is held for five steps, giving a 25\,ms control interval.

\subsection{Tasks}

We use named predicates to keep the specifications compact: $\mu_X$ denotes occupancy of region $X$, and $\mu_{\mathrm{safe}}$ collects the embodiment's safety constraints. Times are in seconds relative to the enclosing operator; all windows are restricted to the episode, and an unsubscripted $\eventually$ ranges over its remaining duration. The episode length of each scenario is chosen such that there is enough time to complete the task.
The tasks are illustrated in Fig.~\ref{fig:pillar_pull_rollout}.

\paragraph{Unlock}
The Crazyflie must stay in A for 0.6\,s to unlock, enter Q within 2\,s of unlocking, then return home H within 10\,s. Q is forbidden before activation:
\begin{equation}
\begin{aligned}
\psi_{\mathrm{unlock}}={}&\always_{[0,0.6]}(\mu_A\wedge\neg\mu_Q)
\wedge\eventually_{[0.6,2.6]}(\mu_Q\wedge\eventually\mu_H),\\
\phi_{\mathrm{unlock}}={}&\always_{[0,10]}\mu_{\mathrm{safe}}
\wedge\bigl(\neg\mu_Q\,\mathbf{U}_{[0,9.4]}\,\psi_{\mathrm{unlock}}\bigr).
\end{aligned}
\end{equation}
Here, safety requires staying within world bounds. The 2\,s entry deadline starts at the end of the dwell.

\paragraph{Random Cue}
A random cue visible for 1\,s selects A or B. The robot must first avoid the selected region, then stay inside it for 0.5\,s by the 5\,s deadline. Let $c_X$ denote the episode's cue chose $X$:
\begin{equation}
\begin{aligned}
\psi_X={}&\always_{[1,2]}\neg\mu_X
\wedge\eventually_{[2,4.5]}\always_{[0,0.5]}\mu_X,\\
\phi_{\mathrm{cue}}={}&\always_{[0,5]}\mu_{\mathrm{safe}}\wedge
\bigwedge_{X\in\{A,B\}}\bigl(c_X\implies\psi_X\bigr).
\end{aligned}
\end{equation}
Safety denotes world bounds for the drone, and workspace, joint-limit, and keepout constraints for the arm.

\paragraph{Ping-pong}
The robot redirects an incoming ball into a target volume. Let $\mu_{\mathrm{hit}}$, $\mu_{\mathrm{out}}$, $\mu_{\mathrm{tar}}$, and $\mu_{\mathrm{play}}$ denote valid incoming contact, sufficient outgoing velocity, ball target occupancy, and ball-in-play conditions. Crazyflie requires a strike and subsequent target entry within 6\,s:
\begin{equation}
\begin{aligned}
\psi_{\mathrm{strike}}={}&\mu_{\mathrm{hit}}\wedge
\eventually_{[\delta,0.08]}\mu_{\mathrm{out}},\\
\phi_{\mathrm{CF\text{-}PP}}={}&\always_{[0,6]}\mu_{\mathrm{safe}}\\
&\wedge\bigl((\mu_{\mathrm{play}}\wedge\neg\mu_{\mathrm{tar}})\,\mathbf{U}_{[0,6]}\\
&\qquad(\psi_{\mathrm{strike}}\wedge
(\mu_{\mathrm{play}}\,\mathbf{U}_{[\delta,6]}\,\mu_{\mathrm{tar}}))\bigr).
\end{aligned}
\end{equation}
Here $\delta=0.01$ is one evaluation step, enforcing post-contact events; safety is drone world bounds. The Panda Ping-pong benchmark scores target entry alone:
\begin{equation}
\phi_{\mathrm{Panda\text{-}PP}}=\eventually_{[0,2.5]}\mu_{\mathrm{tar}}.
\end{equation}

\paragraph{Pillar Pull} Inspired by the game Jenga, this task requires the drone to remove any two of five supporting pillars within 10\,s while preserving the plate and block:
\begin{equation}
\begin{aligned}
\phi_{\mathrm{pull}}={}&\always_{[0,10]}
(\mu_{\mathrm{safe}}\wedge\mu_{\mathrm{plate}}\wedge\mu_{\mathrm{block}})\\
&\wedge\eventually_{[0,10]}
\bigvee_{1\leq i<j\leq5}(\mu_{\mathrm{out},i}\wedge\mu_{\mathrm{out},j}).
\end{aligned}
\end{equation}
The predicates encode drone world bounds, plate height and tilt, block stability and placement, and individual pillar removal. The conjunction inside $\eventually$ requires both pillars to be out simultaneously. This is the hardest task out of all since it does not specify which two pillars to be removed, and can be any combination of 2 out of 5.

\begin{table*}[t]
\centering
\caption{Overall baseline comparison using task success rates (\%, mean $\pm$ sample standard deviation over six training seeds). Evaluation uses 1000 held-out episodes per policy per seed.}
\label{tab:baseline-success}
\footnotesize
\setlength{\tabcolsep}{3pt}
\renewcommand{\arraystretch}{1.12}
\begin{tabular}{@{}lcccc@{}}
\toprule
Task & STL-SVPG P64 & PPO & SAC & TD-MPC2 \\
\midrule
CF Unlock & \textbf{87.67 $\pm$ 20.23} & 0.00 $\pm$ 0.00 & 0.00 $\pm$ 0.00 & 0.02 $\pm$ 0.04 \\
CF Random Cue & 71.93 $\pm$ 2.95 & 13.98 $\pm$ 4.18 & 19.53 $\pm$ 9.72 & \textbf{74.33 $\pm$ 17.14} \\
CF Ping-Pong & \textbf{11.30 $\pm$ 4.36} & 0.00 $\pm$ 0.00 & 0.00 $\pm$ 0.00 & 0.00 $\pm$ 0.00 \\
CF Pillar Pull & \textbf{5.35 $\pm$ 4.09} & 0.00 $\pm$ 0.00 & 0.00 $\pm$ 0.00 & 0.78 $\pm$ 1.56 \\
Panda Random Cue & \textbf{99.98 $\pm$ 0.04} & 83.33 $\pm$ 40.82 & 25.00 $\pm$ 41.83 & 96.72 $\pm$ 2.98 \\
Panda Ping-Pong & \textbf{35.78 $\pm$ 12.99} & 0.00 $\pm$ 0.00 & 6.70 $\pm$ 9.23 & 0.32 $\pm$ 0.38 \\
\bottomrule
\end{tabular}
\vspace{-10pt}
\end{table*}

\paragraph{Branching}
The Crazyflie may follow A$\rightarrow$C or B$\rightarrow$D, reaching the first region by 2\,s and its corresponding goal within another 3\,s:
\begin{equation}
\begin{aligned}
\psi_{AC}={}&\eventually_{[0,2]}(\mu_A\wedge\eventually_{[0,3]}\mu_C)
\wedge\always_{[0,6]}\neg\mu_B,\\
\psi_{BD}={}&\eventually_{[0,2]}(\mu_B\wedge\eventually_{[0,3]}\mu_D)
\wedge\always_{[0,6]}\neg\mu_A,\\
\phi_{\mathrm{branch}}={}&\always_{[0,6]}\mu_{\mathrm{safe}}\wedge(\psi_{AC}\vee\psi_{BD}).
\end{aligned}
\end{equation}
Each route avoids the alternative first region throughout the episode. Safety uses world bounds. The disjunction tests population diversity across valid routes.

\subsection{Evaluation Protocol}
We measure performance by complete task success: a rollout succeeds when the full task specification has positive exact robustness. For each main benchmark, we evaluate the policies from six independent training seeds on a shared set of 1,000 held-out initial conditions. We report the mean success rate and sample standard deviation across training seeds. Policy evaluation uses deterministic actions, while TD-MPC2 uses its evaluation-mode planner.
The Branching study additionally evaluates the policy population to measure coverage of alternative successful routes.

\subsection{Baselines}
We compare STL-SVPG with three reinforcement learning methods: proximal policy optimization (PPO)~\cite{schulman_proximal_2017}; soft actor--critic (SAC)~\cite{haarnoja_soft_2018}; and TD-MPC2~\cite{hansen_td-mpc2_2024}, which learns a dynamics model and uses model-predictive control. In a similar style to prior work on reward engineering~\cite{ma_eureka_2024}, we generate task-specific dense rewards for these baselines with an iterative process leveraging Codex. The resulting dense rewards combine progress shaping, completion bonuses, and control and safety penalties. STL-SVPG instead derives its training objective from the task specification alone. PPO and SAC use interaction budgets matched to the full STL-SVPG population in terms of simulation steps. TD-MPC2 instead matches the training time on GPU and has to use less simulation steps due to the large model it learns with planning in the loop.

\section{Results}
\label{sec:exp}

We aim to answer the following questions:

\begin{enumerate}
    \item Can STL-SVPG learn policies for tasks with extremely sparse success signals?
    \item What does direct STL robustness optimization contribute compared to a dense reward?
    \item How much does maintaining a policy population contribute?
    \item Does STL-SVPG learn multi-modal solutions to the task?
    \item Does a policy learned in simulation transfer to real robots?
\end{enumerate}

\subsection{Learning With Sparse Success Signals}
STL-SVPG delivers the highest mean task success on five of six benchmarks across two robot embodiments, the crazyflie drone (CF) and the Panda arm (Table~\ref{tab:baseline-success}). The two only tasks where baselines achieve comparable performance are the Panda and CF Random Cue. Since they mainly consist of reaching a single goal in time, it is easy to design dense rewards for them. However, for the remaining tasks, where success requires longer time scales and/or interacting with the world with contact, our approach largely outperforms the baselines. On Unlock, it achieves 87.67\% success where PPO and SAC achieve zero and TD-MPC2 reaches 0.02\%. 
On CF Ping-Pong, it is the only method with non-zero success rate.
In CF Pillar Pull, it is the only method to achieve a success rate greater than 1\%, while PPO and SAC completely fail, and TD-MPC2 has very rare success, and not on all seeds.
Finally, it achieves over five times SAC's success on Panda Ping-Pong (35.78\% versus 6.70\%), while PPO completely failed and TD-MPC2 again only had rare successes.
Overall, these results demonstrate that our approach is capable of training policies despite extremely sparse success signals.

\subsection{Benefits of Direct STL Optimization}
We run ablation on the optimization objective using the same SVPG procedure, but switch the gradient used for policy gradient with either automatically differentiated dense rewards, or via finite difference.
Direct STL optimization strengthens learning beyond dense reward shaping (Table~\ref{tab:objective-ablation}). The STL objective nearly doubles Unlock success, from 45.17\% to 87.67\%, and enables Crazyflie Ping-Pong completion where dense return achieves zero. It also improves Random Cue from 66.07\% to 71.93\%. The dense finite difference baseline reaches at most 0.47\%. The consistent gains across tasks show that changing the optimizer alone is insufficient for tasks with sparse success signals. Instead, direct optimization of the trajectory-level STL specification, our core design choice, is essential to align the learning objective with the ordered events and downstream outcomes required for task completion.

\begin{table}[H]
\centering
\vspace{-10pt}
\caption{Objective comparison and black-box dense baseline using task success rates (\%, mean $\pm$ sample standard deviation over six training seeds). Evaluation uses 1000 held-out episodes per policy per seed.}
\label{tab:objective-ablation}
\footnotesize
\setlength{\tabcolsep}{3pt}
\renewcommand{\arraystretch}{1.12}
\begin{tabular}{@{}lccc@{}}
\toprule
Task & STL-SVPG & Dense diff. & Dense FD \\
\midrule
CF Unlock & \textbf{87.67 $\pm$ 20.23} & 45.17 $\pm$ 32.58 & 0.00 $\pm$ 0.00 \\
CF Random Cue & \textbf{71.93 $\pm$ 2.95} & 66.07 $\pm$ 6.29 & 0.47 $\pm$ 0.79 \\
CF Ping-Pong & \textbf{11.30 $\pm$ 4.36} & 0.00 $\pm$ 0.00 & 0.00 $\pm$ 0.00 \\
\bottomrule
\end{tabular}
\end{table}

\subsection{Benefits of Population-Based Learning}
In this ablation, we experiment using only a single particle in the training instead of 64. This is essentially a comparison to directly using gradient descent on the STL gradients.
Maintaining a policy population substantially expands STL-SVPG's learning capability (Table~\ref{tab:population-ablation}). Increasing from one to 64 particles raises mean success on every benchmark, taking Unlock from zero to 87.67\% and Panda Ping-Pong from 7.78\% to 35.78\%. Crazyflie Ping-Pong success increases nearly threefold, while Panda Random Cue reaches near-perfect completion. These results establish the practical value of population-based search: the full method discovers successful behaviors on tasks where the single-policy variant fails and improves performance across both embodiments.

\begin{table}[H]
\centering
    \vspace{-10pt}
\caption{Population ablation with the same per-particle rollout and iteration schedule using task success rates (\%, mean $\pm$ sample standard deviation over six training seeds). Evaluation uses 1000 held-out episodes per policy per seed.}
\label{tab:population-ablation}
\footnotesize
\setlength{\tabcolsep}{3pt}
\renewcommand{\arraystretch}{1.12}
\begin{tabular}{@{}lcc@{}}
\toprule
Task & STL-SVPG P64 & STL-SVPG P1 \\
\midrule
CF Unlock & 87.67 $\pm$ 20.23 & 0.00 $\pm$ 0.00 \\
CF Random Cue & 71.93 $\pm$ 2.95 & 68.63 $\pm$ 9.79 \\
CF Ping-Pong & 11.30 $\pm$ 4.36 & 4.03 $\pm$ 5.00 \\
CF Pillar Pull & 5.35 $\pm$ 4.09 & 0.00 $\pm$ 0.00 \\
Panda Random Cue & 99.98 $\pm$ 0.04 & 93.22 $\pm$ 9.58 \\
Panda Ping-Pong & 35.78 $\pm$ 12.99 & 7.78 $\pm$ 10.46 \\
\bottomrule
\end{tabular}
\end{table}



\subsection{Diverse Successful Policy Populations}
\begin{figure}[t]
    \centering
\includegraphics[width=0.65\linewidth]{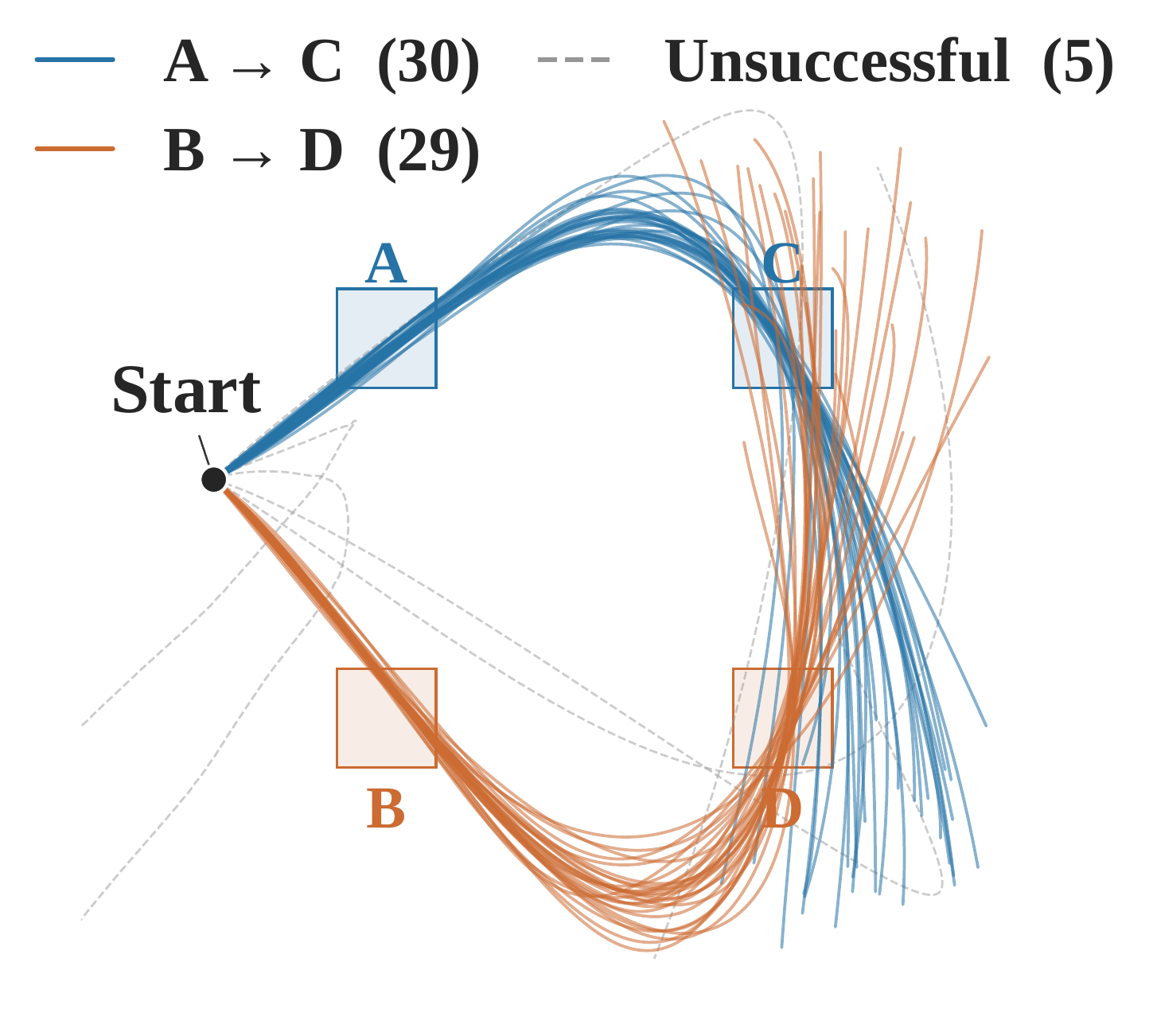}
    \caption{Top-down view of all 64 policies on one reset from a Branching training instance: 30 A--C successes, 29 B--D successes, and five failures.}
    \label{fig:branching}
        \vspace{-10pt}
\end{figure}

We use the Branching task specifically to visualize whether our proposed algorithm learns multiple solutions to the task. STL-SVPG learns a diverse population of successful solutions, retaining both Branching routes in every one of six runs. When evaluated on 128 held-out initial conditions, the mean success rate of the final chosen policy is 99.65\%. When inspecting the behavior of all policies in the population (Figure~\ref{fig:branching}), we see an average split of 34 A--C and 30 B--D policies out of 64 particles. Thus, high task success coexists with distinct, consistently executed strategies.

\begin{figure}[!hbp]
\centering
    \vspace{-20pt}
\includegraphics[width=0.65\linewidth]{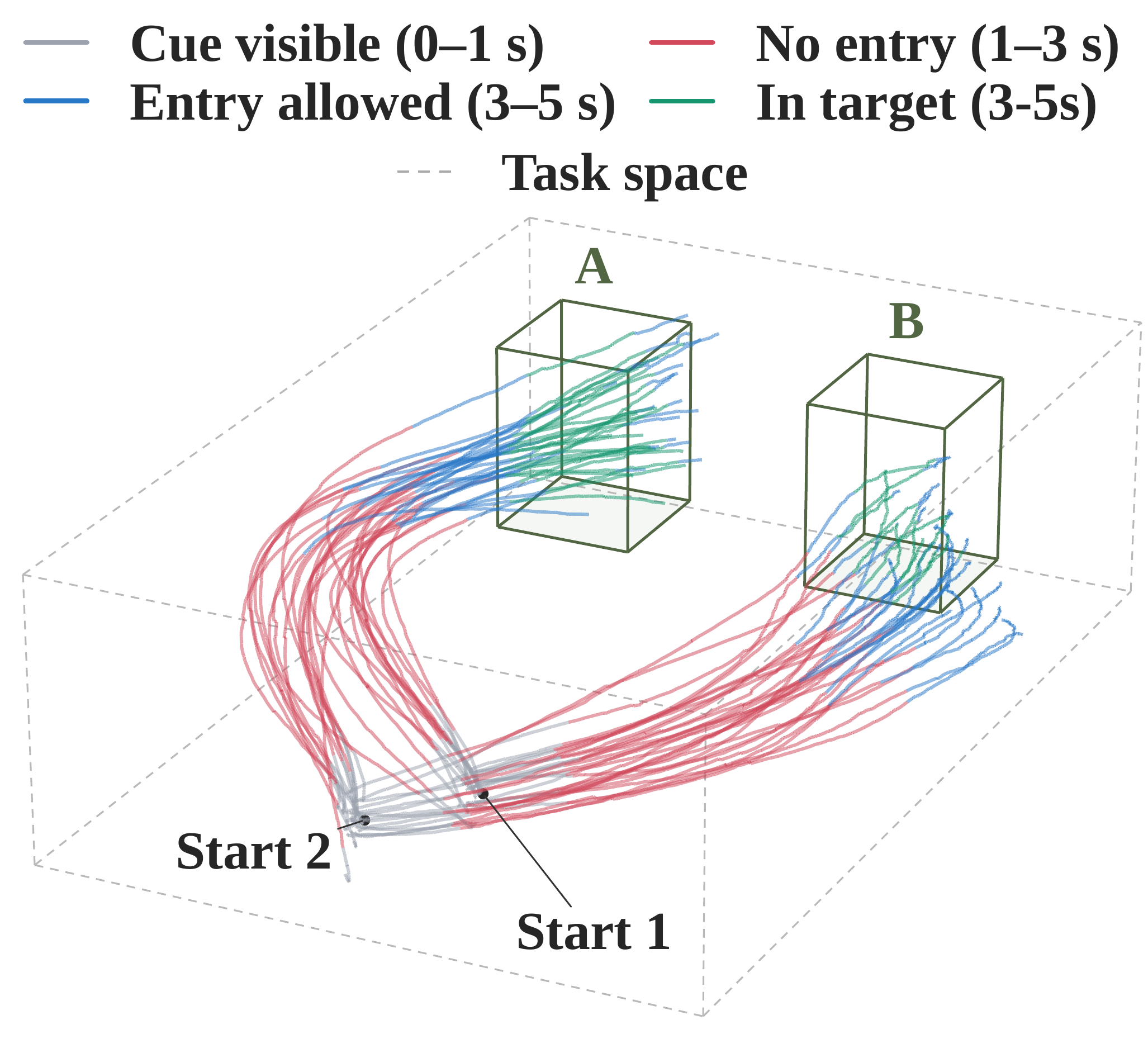}
\caption{Real world policy rollout trajectory of Crazyflie Random Cue from two starting positions. Gray, red, blue, and green portions of the trajectory denotes different phases of the task.}
\label{fig:real-random-cue}
\end{figure}

\subsection{Sim-to-Real Transfer on Crazyflie}
We test Crazyflie Random Cue and Ping-pong on real hardware. Since Ping-pong requires physical contact with the ball, we attach a 3D printed cage around the brushless version of the Crazyflie. We train new deployment policies using new scene and drone cage geometries and brushless motor dynamics.
For Ping-pong, We also change task success to only when the ball enters the target through the top face to simulate a box. These changes significantly decreased simulated success rate in Ping-pong due to more unpredictable contact physics and stricter success requirement.
We use a launcher to throw the ball, and track its trajectory with a Realsense depth camera.

STL-SVPG's learned behavior carries over to physical Crazyflie execution. The simulation-trained Random Cue policy completes 37 of 60 hardware episodes successfully at a success rate of 61.67\%. Figure~\ref{fig:real-random-cue} shows the trajectory executed by the cue-conditioned temporal task on the real robot.
Furthermore, the simulation trained policy transfer to contact in real world. We run the policy for 60 real world episodes.
Compared to 25\% contact rate, 3.91\% valid strike rate, and 0.78\% success rate in simulation, we achieve the 33.33\% contact rate, 6.67\% valid strike rate, and 1.67\% on the real platform. 2 of the real world episodes ends up sending the ball into the target, but in one of those episodes the drone did not recover and crashed after contact. 
An example successful real world rollout is shown in Figure~\ref{fig:real-ping-pong}.


%% file: 5_conclude.tex
\section{Limitations and Conclusion}

STL specifications offer more than a monitor of whether a robot succeeds: their quantitative structure can guide policy learning toward complete task execution even in unsuccessful trials. Our results suggest that combining this feedback with population-based policy optimization is a promising direction for tasks where improving local progress does not necessarily improve overall success.
Nevertheless, this approach depends on informative quantitative specifications and sufficiently accurate differentiable dynamics. Hardware results also reveal a gap between transferring component behaviors and completing the full task: Random Cue achieves temporal-task success, whereas Ping-pong transfers contact and occasional valid strikes but not complete target delivery. Closing this gap will require better treatment of contact-model uncertainty, better domain randomization, and more reliable credit assignment across successive task events during training. These challenges motivate future work in STL-guided policy-learning algorithms that retain performance under real-world uncertainty.

\section*{ACKNOWLEDGMENT}
OpenAI Codex was used for improving clarity on non-technical sections of the manuscript and assist in implementing experimental code and plotting. All technical content, experimental design, results, and conclusions were developed, verified, and carefully vetted by the authors.